%% file: main.tex
\documentclass[sigconf]{acmart}
\makeatletter
\def\@ACM@checkaffil{
    \if@ACM@instpresent\else
    \ClassWarningNoLine{\@classname}{No institution present for an affiliation}%
    \fi
    \if@ACM@citypresent\else
    \ClassWarningNoLine{\@classname}{No city present for an affiliation}%
    \fi
    \if@ACM@countrypresent\else
        \ClassWarningNoLine{\@classname}{No country present for an affiliation}%
    \fi
}
\makeatother

\AtBeginDocument{%
  \providecommand\BibTeX{{%
    \normalfont B\kern-0.5em{\scshape i\kern-0.25em b}\kern-0.8em\TeX}}}
\pdfoutput=1
\usepackage{booktabs} 
\usepackage[normalem]{ulem}
\usepackage{url}
\usepackage{xcolor} 
\usepackage{algorithm}
\usepackage{algpseudocode}
\usepackage{amsmath}
\usepackage{enumerate}
\usepackage[shortlabels]{enumitem}
\usepackage[mathscr]{eucal}
\usepackage{graphicx}
\usepackage{subfigure}
\usepackage{makecell}
\usepackage{mathtools}
\usepackage{multirow}
\usepackage[font=rm]{caption}
\usepackage{shortvrb}
\usepackage{tabularx}
\usepackage{verbatim}
\usepackage{xspace}
\usepackage{listings}
\usepackage{fontawesome}
\usepackage[multiple]{footmisc}
\usepackage[all]{nowidow}
\usepackage{balance}
\usepackage{setspace} 
\usepackage{hyperref}
\hypersetup{colorlinks,allcolors=black}
\pdfoutput=1

\newcommand{\eat}[1]{}

\input{localmacros.tex}

\usepackage[medium,compact]{titlesec}

\setcopyright{none}

\begin{document}
\pdfoutput=1

\title[IntentDial: An Intent Graph based Multi-Turn Dialogue System]{IntentDial: An Intent Graph based Multi-Turn Dialogue System with Reasoning Path Visualization}

\author{Zengguang Hao}
\affiliation{%
  \institution{Ant Financial Services Group}
}

\author{Jie Zhang}
\affiliation{%
  \institution{Ant Financial Services Group}
}
 
\author{Binxia Xu}
\affiliation{%
  \institution{Ant Financial Services Group}
}

\author{Yafang Wang}
\authornote{Corresponding author; Email: yafang.wyf@antfin.com}
\affiliation{%
 \institution{Ant Financial Services Group}
}

\author{Gerard de Melo}
\affiliation{%
  \institution{HPI / University of Potsdam}
}

\author{Xiaolong Li}
\affiliation{
\institution{Ant Financial Services Group}
}

\renewcommand{\shortauthors}{Hao et al.}

\begin{abstract}
Intent detection and identification from multi-turn dialogue has become a widely explored technique in conversational agents, for example, voice assistants and intelligent customer services. The conventional approaches typically cast the intent mining process as a classification task. Although neural classifiers have proven adept at such classification tasks, the \emph{black box} issue of neural network models often impedes their practical deployment in real-world settings. We present a novel graph-based multi-turn dialogue system called \emph{IntentDial}, which identifies a user's intent by identifying intent elements and a standard query from a dynamically constructed and extensible intent graph using reinforcement learning.
In addition, we provide visualization components to monitor the immediate reasoning path for each turn of a dialogue, which greatly facilitates  further improvement of the system.
\end{abstract}


\maketitle

\input{Intro.tex}
\input{system.tex}

\input{demonstration.tex}
\input{conclu.tex}

\small
\bibliographystyle{ACM-Reference-Format}
\bibliography{main}

\end{document}

%% file: localmacros.tex
\newcommand{\Patk}[1]{\mbox{\Pat@$#1$}}
\newcommand{\RBPatp}[1]{\mbox{\RBP@$#1$}}

\newcommand{\NDCGatk}[1]{\mbox{\NDCG@$#1$}}
\newcommand{\ERRatk}[1]{\mbox{\ERR@$#1$}}

\newcommand{\myurl}[1]{{\url{#1}}}

\newcommand{\myparagraph}[1]{\vspace{0.2\baselineskip}\noindent{\textbf{#1}}.~}

\newcommand{\mycomment}[1]{}

\newlength{\onedigit}
\newcounter{todocount}

%% file: intro.tex
\section{Introduction}
An accurate customer intent identification system is crucial for both customers and enterprises. An intelligent customer service that is able to effectively detect user intents can greatly accelerate the process of dealing with their problems. Within enterprises, if one is able to precisely pinpoint a customer’s issue, the corresponding staff and resources can be allocated immediately, leading not only to a reduction in cost and resources but also improving the customer’s satisfaction and sentiment with regard to the enterprise.
For a more natural interaction, the intent recognition procedure is often designed as a dialogue system. For instance, the PARLANCE system~\cite{hastie2013demonstration} provides an interactive search procedure for users and allows for processing smaller \emph{chunks} of user input instead of integrated utterances; Goo et al.~\cite{goo2018slot} propose a slot-gated model to infer the user's intent from utterances, validating its effectiveness on the ATIS and Snips datasets; Li et al. \cite{li-etal-2017-end} construct an end-to-end task-oriented neural dialogue system combining supervised learning and reinforcement learning. 
Apart from this, conventional approaches for intent detection also often regard it as a classification task, in which user utterances are labelled with a corresponding intent or a standard query, and an intent classifier is trained on this data \cite{kathuria2010classifying,caruccio2015understanding,purohit2015intent}. While such intent classification models have proven effective in both academia as well as industry, they also have notable shortcomings. 
First, whenever the business scope of a company expands or shifts, new intent labels or standard queries need to be added, and the model needs to be retrained. Second, state-of-the art deep neural classifiers are \emph{black boxes} in the sense that they lack interpretability, making it non-trivial to analyze and improve the system. 

In this work, we instead propose a multi-turn dialogue system, \textbf{IntentDial}, that identifies the user intent by identifying paths in a custom graph. Our approach constructs such a graph on the fly for the specific user, and learns to identify suitable paths in such graphs that lead to a specific intent.
For the latter, we rely on reinforcement learning (RL), inspired by recent results on using RL for reasoning in knowledge graphs (KGs)  \cite{xiong2017deeppath,das2017go,lin2018multi,fu2019collaborative}. Compared with those current works, our system allows an intermediate state of the reasoning path, which can be more flexible and achieve a more natural interaction process with users.

Given our interpretable inference approach, we further design a reasoning path monitoring platform that visualizes the decision making procedure of the system. This allows us to inspect and consider possible future improvements of the system. For instance, if a large number of incorrect paths pass by the same node in a graph, one can further investigate the role of this node and enhance the performance of the system. We have successfully trained a dialogue system based on this framework and applied it to the intelligent customer services in several business scenarios of Ant Financial Services Group.

The remainder of our paper is organized as followed: Section 2 describes the architecture and methods adopted in our system. Section 3 presents the user interface based on two real-world use cases. Section 4 concludes the paper. 
A screencast of our system is available at: \url{https://youtu.be/tozgo3KZFgA}

%% file: system.tex
\section{System Overview}
\begin{figure*}[htbp]
    \flushleft
    \begin{minipage}[t]{0.56\linewidth}
    \flushleft
    \includegraphics[width=\textwidth]{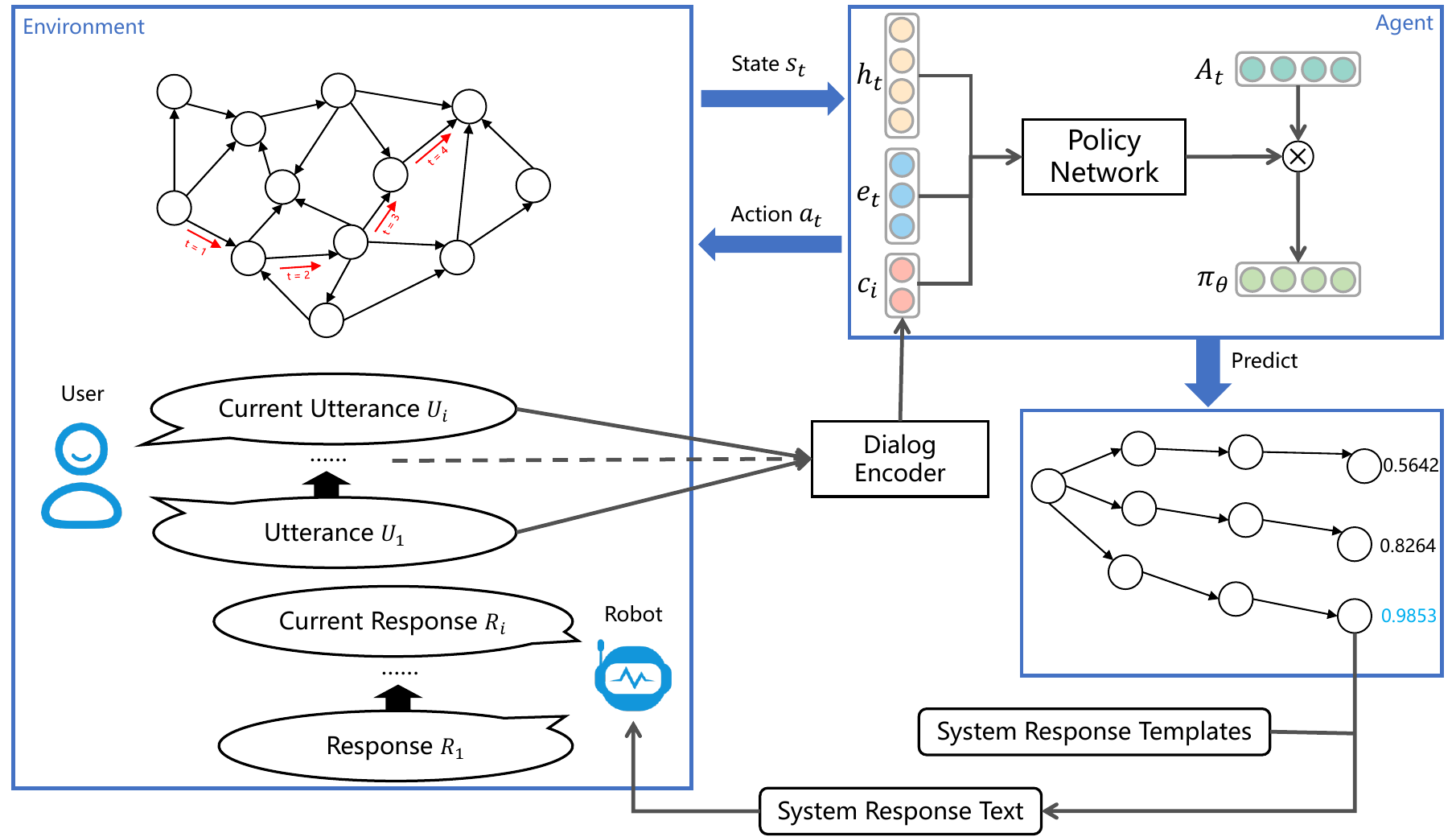} 
    \caption{System Overview} 
    \label{fig:system_overview}
    \vspace{-12pt}
    \end{minipage}%
    \hspace{5pt}
    \begin{minipage}[t]{0.4\linewidth} 
    \centering
    \includegraphics[width=\textwidth, height=5.5cm]{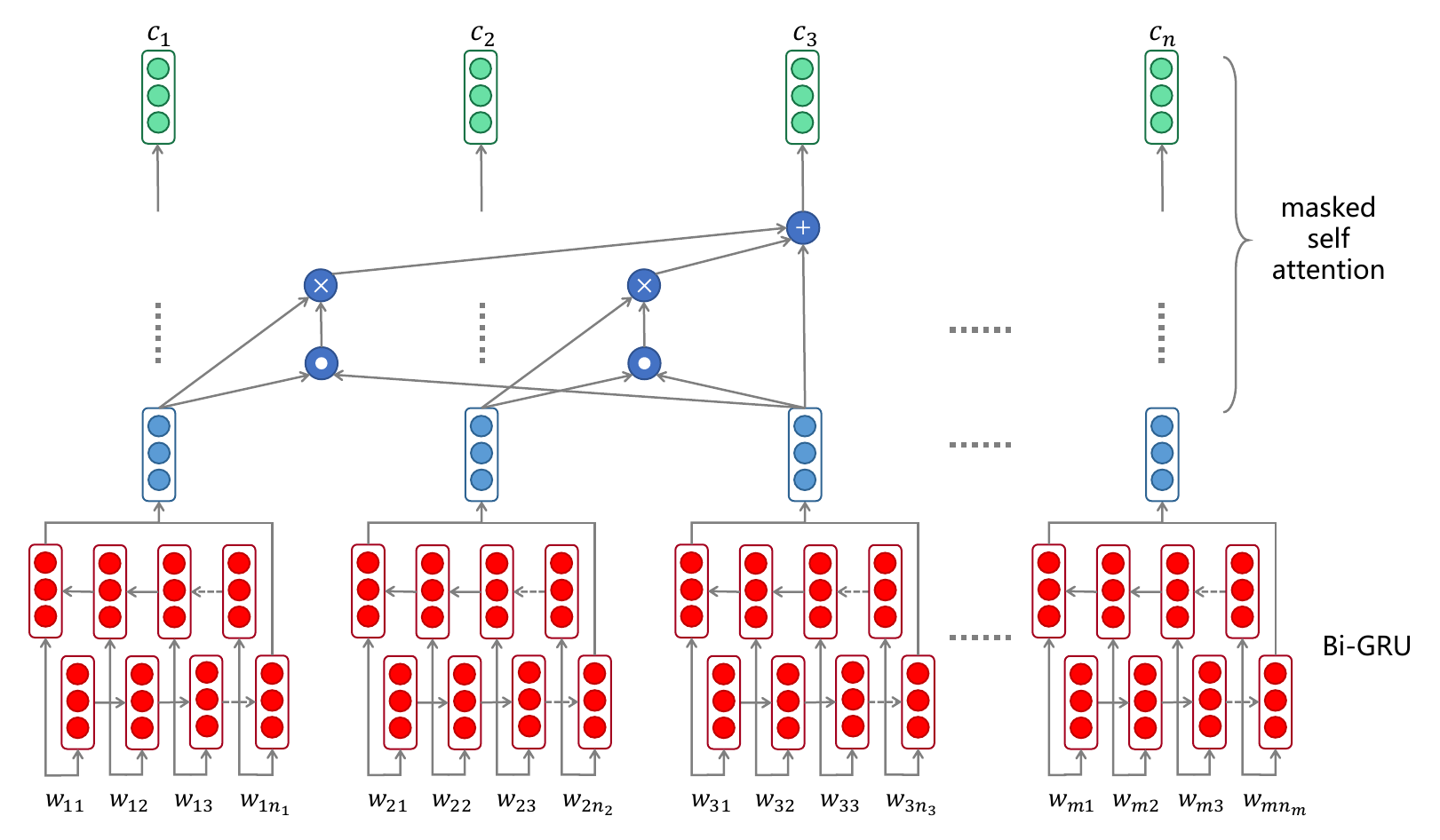}
    \caption{Structure of the dialogue encoder.}
    \label{fig:main_model}
    \vspace{-10pt}
    \end{minipage}
\end{figure*}
Our system supports inputs both in the form of written and spoken language.
The primary objective of the system is to consider user utterances from the dialogue input and identify the underlying user intent.
To achieve this, it constructs a graph and then relies on reinforcement learning to infer a reasonable path that points to a specific intent. Figure \ref{fig:system_overview} illustrates the overall architecture of the system. At the $i$-th round of dialogue, the system first encodes the current utterance $U_i$ along with previous historical utterances $U_1$ to $U_{i-1}$ as a context embedding vector $c_i$. The next step is to draw on  reinforcement learning to find a suitable path that starts at a specified starting node to a suitable end node, which corresponds to a specific intent. In the path finding process, at step $t$, the agent acquires the state $s_t$, which includes the current entity $e_t$ in the graph, the root node $e_r$ of the graph, and the context embedding $c_i$, and encodes the path history $(e_r, r_1, e_1, \dots, r_t, e_t)$ as $h_t$. Subsequently, $h_t$, $e_t$, and $c_i$ are concatenated as the input to the policy network, which integrates the action space $A_t$ and emits an action policy $\pi_\theta$ representing the action probabilities. We run a number of trajectories in the prediction stage to locate the most suitable path among them. Finally, depending on the identified end entity, we select the corresponding response template and fill it to obtain the system response $R_t$ for the user.

\vspace{-10pt}
\subsection{Knowledge Graph}
\label{sec:kg}
The knowledge graph $\mathcal{G} = \{(e_s, r, e_o) \mid e_s, e_o \in \mathcal{E}, r \in \mathcal{R}\}$ consists of three distinct kinds of nodes, i.e., query nodes $E_q = \{e_{q_i} \mid 1 \le i \le n_q\}$, feature nodes $E_f = \{e_{f_i} \mid 1 \le i \le n_f\}$, and a single root node $e_r$. A triple $(e_s, r, e_o)$ denotes an edge $e_s \stackrel{r}{\longrightarrow} e_o$ in the knowledge graph. $\mathcal{E}$ is the entity set and  $\mathcal{R}$ the relation set. Each query node stands for a query in the query base, indicating the intent of the user. The main goal in our system is to find a query according with the user's intent. For this reason, the out-degree of query nodes is 0. The feature nodes are manually labeled from the query text in the query base and represent specific features of queries. Certain special feature nodes are essential for path inference and are thus considered \emph{key nodes}, which indicate the sub-intents of the user. The knowledge graph is designed to start from one specific kind of key node. For convenience, the root node is added and only points to this kind of key node. Hence, the agent may start from the root node to find paths in the path reasoning process.

\subsection{Dialogue Setting}
The goal in the task-oriented dialogue is to obtain the user intent according to the user utterances. As mentioned above, the query stands for the intent and also determines the response action for the dialogue. However, in a particular dialogue, we may lack information extracted from the utterances in the initial stage, preventing us from reaching any query node during path inference. To address this, we introduce the key features as response actions, and also design a response template for them. If the result path ends at a key node, the system will instead ask the user about other key nodes that are missing in the path.

\subsection{Dialogue Encoder}
\label{sec:Dialogue_encoder}
As depicted in Figure \ref{fig:main_model}, the multi-turn dialogue encoder consists of 3 layers: the embedding layer, sentence encoding layer, and context encoding layer. The input data is a batch of multi-turn dialogues. 

\myparagraph{Sentence Encoding Layer}
The encoding layer projects the sentence $U_i$ for each round in a dialogue to a vector. The input sequence of tokens $U_i = \{w_{i1}, w_{i2}, \dots, w_{in_i}\}$ is passed through a Bi-GRU (Bidirectional Gated Recurrent Unit)~\cite{cho2014learning}. We concatenate the final two bidirectional unit states to generate the sentence embedding.

\begin{figure*}[htbp]
    \centering
    \includegraphics[width=0.72\textwidth]{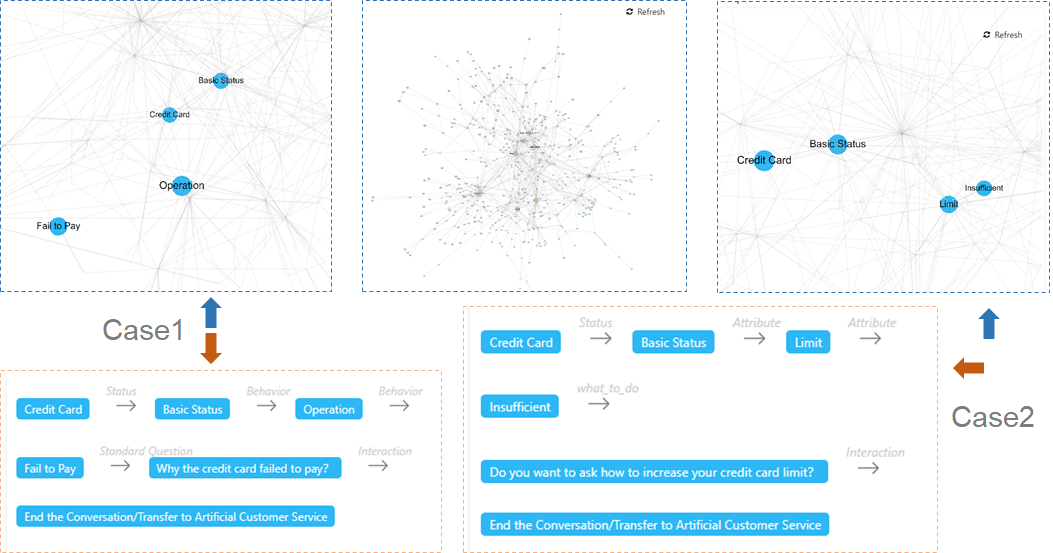}
    \caption{Visualization of reasoning paths -- The graph in the middle is the visualization of the knowledge base we rely on in our system.}
    \label{visualization}
    \vspace{-0.1in}
\end{figure*}

\myparagraph{Context Encoding Layer}
This layer integrates forward turns of sentences in a dialogue and emits a context embedding $c_i$ in each turn step. In each round, we also need to consider the contents of previous turns of sentences and extract useful information that can be imported into the context embedding. Hence, the model incorporates a Masked Self-Attention mechanism (dot-product attention)~\cite{vaswani2017attention} that masks the backward turns of dialogue sentences.

\subsection{Path Inference}
The system traverses the graph via reinforcement learning in order to find a suitable path that points to the underlying intent. For this, in the $i$-th  round of dialogue, given the current context embedding $c_i$ and knowledge graph $\mathcal{G}$, the path reasoner aims at inferring a path from the root node to some entity.

\subsubsection{Reinforcement Learning Formulation}
The path inference procedure can be viewed as a Markov Decision Process (MDP). Optimization is achieved using REINFORCE~\cite{williams1992simple}, which is similar to MINERVA~\cite{DasDZVDKSM18} and defined as follows:

\myparagraph{States}
For path-based reasoning, at step $t$, we want the state to encode the current observed environment, i.e., $s_t = (e_t, e_r, c_i) \in \mathcal{S}$, where $e_t \in \mathcal{E}$ is the entity in the knowledge graph visited at step $t$. $e_r$ and $c_i$ can be shared by all states.

\myparagraph{Actions}
At step $t$, the reasoner picks one out of all available out-edges of the current entity $e_t$ as its next action $a_t$. Specifically, the set of actions of $e_t$ is 
$A_t = \{(r', e') \mid (e_t, r', e') \in \mathcal{G}\}$, where $(e_t, r', e') \in \mathcal{G}$ is an edge in the knowledge graph $\mathcal{G}$.

\myparagraph{Transitions}
The transition function $f$: $\mathcal{S} \times \mathcal{R} \rightarrow \mathcal{S}$ is formulated as $f(s_t, a_t) = (e_{t + 1}, e_r, c_i)$.

\myparagraph{Rewards}
During training, the model acquires a positive reward +1 if the path arrives at the target entity, and a negative reward 0 otherwise. To obtain a suitable path, if the path passes along key nodes of the target query entity, we also add immediate positive rewards to the steps corresponding to these key nodes.

\subsubsection{Policy Network}
The policy network is as well similar to MINERVA~\cite{DasDZVDKSM18}. First, we leverage a Long Short-Term Memory (LSTM) layer~\cite{hochreiter1997long} to encode the history $H_t = (e_r, r_1, e_1, \dots, r_t, e_t)$ as a history embedding $h_t$.
\begin{equation}
    h_t = \text{LSTM}(h_{t-1}, a_{t-1})
\end{equation}
Then, $h_t$ serves as the input of the policy network $\pi$, which is defined as:
\begin{equation}
    \pi_\theta(a_t \mid s_t) = \text{softmax}(\mathbf{A}_t(\mathbf{W}_2\text{ReLU}(\mathbf{W}_1[h_t;e_t;c_i]))),
\end{equation} 
where $\mathbf{A}_t$ denotes the action space generated by stacking the embeddings of all actions in $A_t$.

\subsection{Training}
The parameters $\theta$ for the entire model are trained by maximizing the expected reward:
\begin{equation}
    \text{J}(\theta) = \mathbb{E}_{D_i \in \mathcal{D}} \mathbb{E}_{c_i \in D_i} \mathbb{E}_{a_1, \dots, a_T \sim \pi_\theta} [R(s_T \mid e_r, c_i)],
\end{equation}
where $D_i = \{c_1, c_2, \dots, c_I\}$ represents a dialogue from dialogue set $\mathcal{D}$.
The parameters $\theta$ are updated based on the following gradient:
\begin{equation}
    \nabla J(\theta) = \nabla_\theta \sum_t R(s_T \mid e_r, c_i) \log \pi_\theta(a_t \mid s_t)
\end{equation}

%% file: demonstration.tex
\section{Case Studies}
\vspace{-2ex}
\begin{figure}[htbp]
    \centering
    \includegraphics[width=0.32\textwidth]{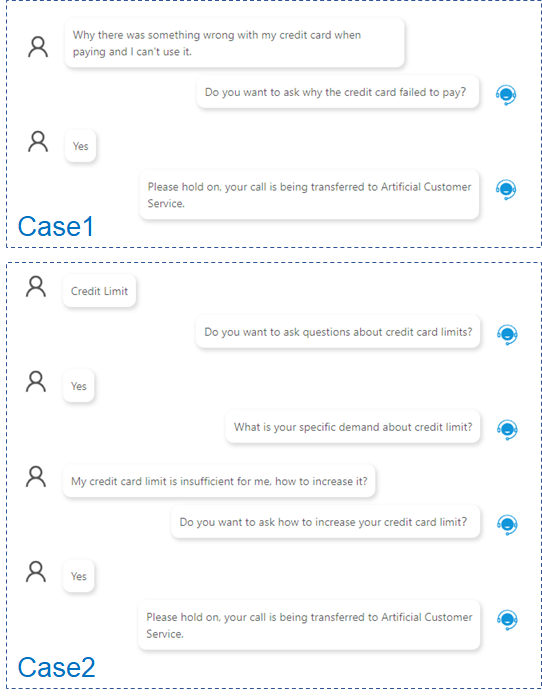}
    \vspace{-1ex}
    \caption{Sample conversations with our system.}
    \label{dialogue}
\end{figure}

The goals of our system are two-fold: 1) We seek a dialogue system that is able to detect a customer's intent and match it with a standard query in our repository so that an accurate and appropriate form of artificial customer service can be provided; 2) We further incorporate visualization components that facilitate the analysis of the system's reasoning process. The user interface of our system consists of two main parts: a chat window and a monitor platform. 
The chat window embedded in the system allows users to initiate a conversation with the intelligent customer service agent. According to the user's description, the agent will either confirm its understanding of the query with the user or request more specific information regarding their demands (Figure~\ref{dialogue}). The visualization component for monitoring the system is presented in Figure~\ref{visualization}.  It provides two forms of visualization: an intent graph with nodes highlighted and a reasoning path resulting from the current intent mining process. 
To better illustrate how the system operates, we next consider two different examples addressing different scenarios: standard query matching and intent elements matching.

\myparagraph{Standard query matching} As presented in Figure~\ref{dialogue} (top), standard query matching is triggered when the overall intent of the user can be captured by matching it directly with an existing standard query. In Case 1, for example, the user's question is ``why there was something wrong with my credit card when paying and I can't use it", from which a reasoning path leading to the standard question ``Do you want to ask why the credit card failed to pay?" can be found (Figure~\ref{visualization} left). Once the query is reconfirmed by the user, the dialogue is transferred to the corresponding artificial customer service agent. 

\begin{figure}[htbp]
    \centering
    \includegraphics[width=0.3\textwidth]{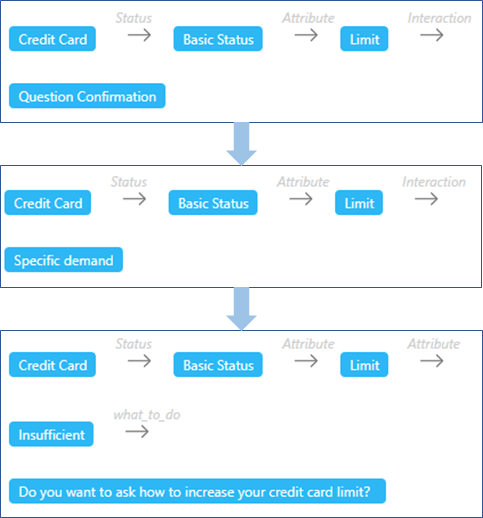}
    \vspace{-1ex}
    \caption{Change of reasoning path during several rounds of a dialogue.}
    \label{step}
\end{figure}

\vspace{-2ex}
\myparagraph{Intent element matching} Although accurately capturing a customer's intent from just a single round of a dialogue is desirable, in real-world settings, due to the ambiguity and complexity of user questions, it is often more feasible to progressively determine the user's underlying intent across several rounds of a dialogue. Figure~\ref{dialogue} (bottom) shows a scenario in which the user's intent does not immediately correspond to a standard query. In the first turn of the dialogue, the user only mentions a part of the intent: ``Credit Limit", so the tail node of the reasoning path based on the provided information is not a standard question as presented in the top of Figure~\ref{step}. As the intent elements ``Credit Card", ``Limit" are confirmed by the user, the agent retrieves a template question so as to guide the user to providing more intent elements, in this case, ``Specific demand". In the meantime, the reasoning path for this round of the dialogue changes (Figure~\ref{step} middle). Subsequently, the user provides further details by stating that ``My credit card limit is insufficient for me, how to increase it?" and a reasoning path can then be placed that leads to the corresponding standard query (Figure~\ref{step} bottom). 
The visualization of the final reasoning results for this dialogue are shown as Case 2 in Figure~\ref{visualization}. In this manner, even if the matching process requires several rounds of dialogue, the form of interaction is very natural and acceptable.

%% file: conclu.tex
\section{Conclusion}
This work showcases IntentDial, a system that is able to detect a user's intent in multi-turn dialogue. The system constructs a custom knowledge graph and relies on reinforcement learning to identify reasoning paths in this graph, which can be visualized to enable interpretability of the inner workings of the inference process. The system not only allows users to experience a much more natural form of interaction with intelligent customer service systems by progressively guiding them to a standard query, but also provides a monitoring platform to guide further improvements of the system. At present, our system only supports reasoning path visualization for a single multi-turn dialogue, which in future work could be expanded by adding an analysis platform for large datasets.